%% file: root.tex
\documentclass[letterpaper, 10 pt, conference]{ieeeconf}

\IEEEoverridecommandlockouts

\usepackage[normalem]{ulem}
\usepackage{xcolor}
\usepackage{graphicx}
\usepackage{subcaption}

\usepackage{amsmath}
\usepackage{amssymb}

\usepackage[bookmarks=true]{hyperref}

\usepackage[capitalise]{cleveref}
\usepackage{enumerate}
\usepackage{balance}

\usepackage[noend]{algpseudocode}
\usepackage{algorithm}
\algrenewcommand\algorithmicrequire{\textbf{Input:}}
\algnewcommand\algorithmicnot{\textbf{not}}
\algdef{SE}[IF]{IfNot}{EndIf}[1]{\algorithmicif\ \algorithmicnot\ #1\ \algorithmicthen}{\algorithmicend\ \algorithmicif}%
\algtext*{EndIf}
\usepackage{varwidth}

\newcommand{\X}{\mathcal{X}}
\newcommand{\Q}{\mathcal{Q}}
\newcommand{\T}{\mathcal{T}}
\newcommand{\ptc}{\Call{ptc}{}}

\def\Xgoal{\X_{\text{goal}}}
\def\Xfree{\X_{\text{free}}}
\def\Qgoal{\Q_{\text{goal}}}
\def\Tgoal{\T_{\text{goal}}}

\def\xstart{x_{\text{start}}}
\def\xrand{x_{\text{rand}}}

\def\xnew{x_{\text{new}}}
\def\xnear{x_{\text{near}}}
\def\tstart{t_{\text{start}}}
\def\qstart{q_{\text{start}}}
\def\tmin{t_{\text{min}}}
\def\tmax{t_{\text{max}}}
\def\tlb{t_{\text{lb}}}
\def\tub{t_{\text{ub}}}
\def\vmax{v_{\text{max}}}

\def\tr{B.\textit{timeRange}}
\def\ntr{B.\textit{newTimeRange}}
\def\bs{B.\textit{batchSize}}
\def\rf{P.\textit{rangeFactor}}
\def\sr{P.\textit{sampleRatio}}
\def\ibs{P.\textit{initialBatchSize}}
\def\bp{B.\textit{batchProbability}}
\def\sib{B.\textit{samplesInBatch}}
\def\ts{B.\textit{totalSamples}}
\def\g{B.\textit{goals}}
\def\ng{B.\textit{newGoals}}
\def\;{;$ \hspace{0.5mm} $}
\def\pgoal{p_{\text{goal}}}

\def\Treegoal{T_{\text{goal}}}

\newcommand{\pluseq}{\mathrel{+}=}

\newcommand{\asteq}{\mathrel{*}=}

\definecolor{cblue}{HTML}{648FFF}
\definecolor{cred}{HTML}{DC267F}
\definecolor{corange}{HTML}{FE6100}
\definecolor{cyellow}{HTML}{FFB000}

\newcommand{\sqboxs}{1.5ex}
\newcommand{\sqbox}[1]{\textcolor{#1}{\rule{\sqboxs}{\sqboxs}}}

\def\funcLowerBoundArrivalTime{LowerBoundArrivalTime}

\begin{document}
%
\title{\LARGE \bf ST-RRT*: Asymptotically-Optimal Bidirectional \\Motion Planning through Space-Time}

\author{Francesco Grothe$^1$, Valentin N. Hartmann$^{1, 2}$, Andreas Orthey$^{1}$, Marc Toussaint$^1$%
\thanks{The research has been supported by the Deutsche Forschungsgemeinschaft (DFG, German Research Foundation) under Germany's Excellence Strategy -- EXC 2120/1 -- 390831618 ``IntCDC''.}%
\thanks{$^{1}$Learning and Intelligent Systems Group, TU Berlin, Germany}%
\thanks{$^{2}$Machine Learning \& Robotics Lab, University of Stuttgart, Germany
        {\tt\footnotesize valentin.hartmann@ipvs.uni-stuttgart.de}
        }%
}

\maketitle

\begin{abstract}
\input{sections/0_abstract}
\end{abstract}


\IEEEpeerreviewmaketitle

\section{Introduction}
\input{sections/1_introduction}

\section{Related Work}\label{sec:related_work}
\input{sections/2_related_work}

\section{The Space-Time RRT* Algorithm}\label{sec:method}

\input{sections/3_algorithm}

\section{Evaluation}\label{sec:experiments}
\input{sections/4_evaluation}

\section{Conclusion}\label{sec:conclusion}
\input{sections/5_conclusion}

\bibliographystyle{IEEEtran}
\balance
\bibliography{IEEEabrv, lit.bib}

\end{document}

%% file: sections/0_abstract.tex
We present a motion planner for planning through space-time with dynamic obstacles, velocity constraints, and unknown arrival time.
Our algorithm, Space-Time RRT* (ST-RRT*), is a probabilistically complete, bidirectional motion planning algorithm, which is asymptotically optimal with respect to the shortest arrival time.
We experimentally evaluate ST-RRT* in both abstract (2D disk, 8D disk in cluttered spaces, and on a narrow passage problem), and simulated robotic path planning problems (sequential planning of 8DoF mobile robots, and 7DoF robotic arms).
The proposed planner outperforms RRT-Connect and RRT* on both initial solution time, and attained final solution cost. 
The code for ST-RRT* is available in the Open Motion Planning Library (OMPL). 

%% file: sections/1_introduction.tex
Motion planning is a fundamental challenge in robotics~\cite{springer_roboticsHandbook}.
In many real-world applications, 
obstacles change positions over time and goals are only valid at specific times.
For applications such as multi-robot assembly, multiple motion scheduling subproblems need to be solved \cite{Hartmann2021LongHorizonMR}. 
Assuming that obstacle trajectories are given a priori, the subproblems can be modelled as \emph{navigation through dynamic environments}. 
Mathematically, this is formulated as planning through a space-time state space.

Efficient and optimal planning through space-time raises three fundamental challenges.
First, since goal arrival times are unknown upfront, it becomes difficult, yet crucial, to define and adjust the time range in a coordinated and meaningful way. 
The second challenge is the representation of kinodynamic constraints in the planning model.
Whether a movement is possible depends on kinematic parameters, velocity, and acceleration.
Lastly, robots should minimize arrival time. Arrival time is crucial for long-horizon planning problems, where optimization of intermediate arrival times is one of the central challenges \cite{Hartmann2021LongHorizonMR}.
These challenges make planning through space-time a demanding problem. 
We are not aware of any sampling-based method which either operates in unbounded space-time or is asymptotically optimal with respect to shortest arrival time.

\begin{figure}[t]
    \centering
    \includegraphics[width=\linewidth,height=0.9\linewidth]{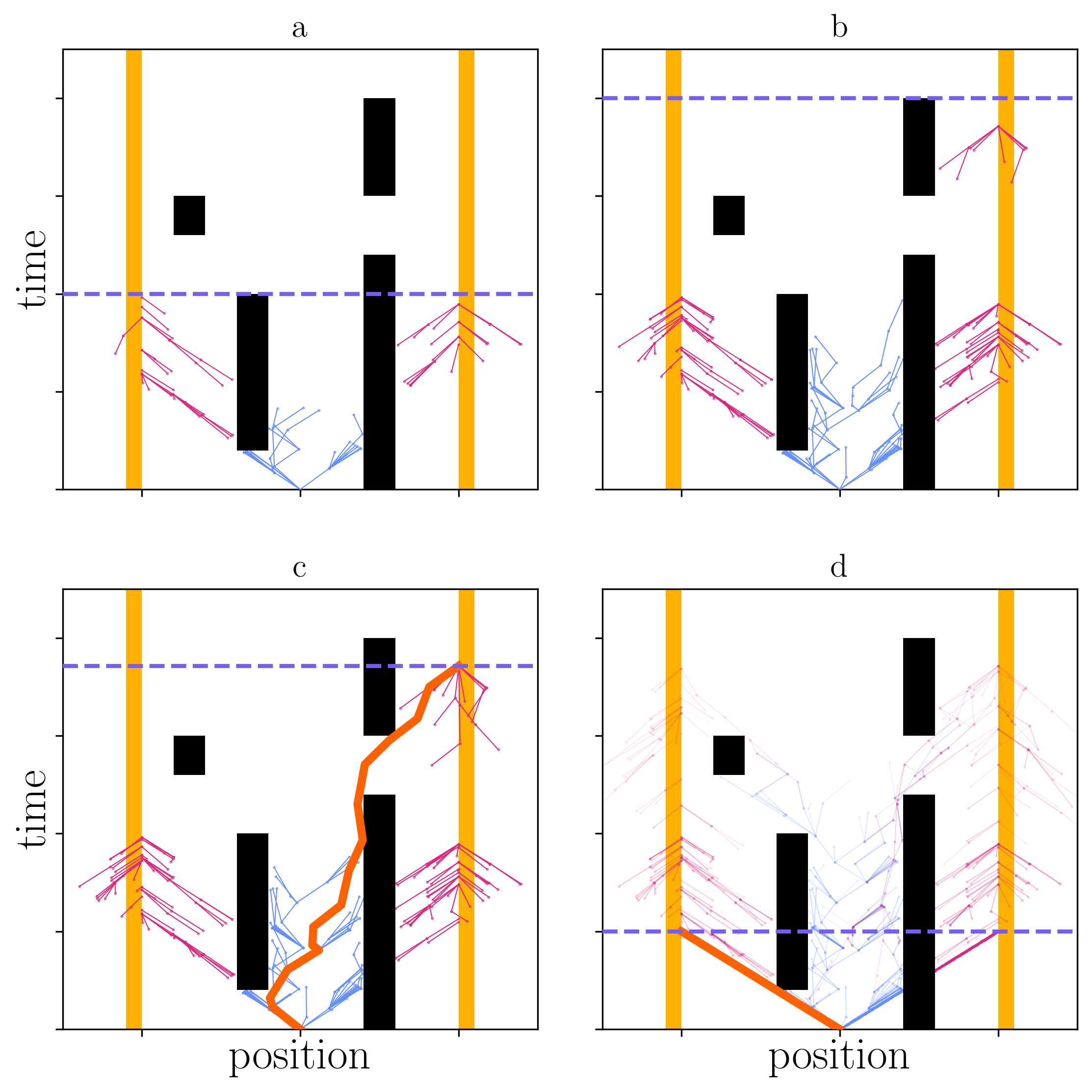}
    \caption{Four snapshots of ST-RRT* in $\mathbb{R}^{1+1}$ (one space, plus one time dimension).
    The forward tree is blue, the backward trees are red, obstacles are black, and the goal regions are yellow.
    (\textbf{a}) Using the initial batch of samples, no solution was found. 
    (\textbf{b}) The upper bound of the time space (dashed line) is expanded, more goal nodes are sampled and the trees are grown. 
    (\textbf{c}) An initial solution is found (orange), and the upper bound is decreased accordingly. 
    (\textbf{d}) Parts of the trees that can not contribute to the solution anymore are pruned (lower opacity), and the final solution after convergence.}
    \label{fig:intro}
\end{figure}

To address those challenges, we develop Space-Time RRT* (ST-RRT*).
The basic operating principle of ST-RRT* is illustrated in \cref{fig:intro}: (\textbf{a}) We compute an initial estimate of a feasible goal time (blue dashed line), and grow both a forward tree from the start state (blue), and a set of reverse trees from the goal regions (red).
If no solution is found given a certain number of samples, the upper time limit in which we generate samples is increased (\textbf{b}).
If a solution (orange) is found (\textbf{c}), the parts of the trees that can not lead to an improved solution are pruned.
This process continues to improve the solution path, and tighten the upper time bound until a termination condition is reached (\textbf{d}).

ST-RRT* is a bidirectional motion planning algorithm that is probabilistically complete and asymptotically optimal with respect to shortest arrival time.
ST-RRT* is able to operate in unbounded time spaces and model velocity constraints.
ST-RRT* is inspired by RRT-Connect \cite{kuffner2000rrt-connect}, with three changed components to attain the stated qualities in space-time. 
Our main contributions are:

\begin{itemize}[]
    \item \textit{Progressive Goal Region Expansion}: ST-RRT* gradually increases the sampled time range to efficiently operate in unbounded time spaces.
    Simultaneously, we adjust the sampling densities over the time dimension to ensure a more uniform sampling distribution.

    \item \textit{Conditional Sampling}: 
    We develop a novel sampling method that prevents the sampling of states which cannot be part of a solution path due to velocity constraints.

    \item \textit{Simplified Rewiring}: 
    To obtain optimal solutions, states are rewired similar to RRT* \cite{karaman2011sampling}.
    In contrast to RRT*, we perform a simplified rewiring step, where only nodes in the set of goal trees are rewired.
\end{itemize}

We demonstrate our algorithm on both abstract (planning for a disk in up to $\mathbb{R}^{8+1}$), and simulated robotic motion planning problems (both robotic arms and mobile robots).

%% file: sections/2_related_work.tex

In the following two sections, we review literature on planning in dynamic environments and time-optimal path planning.
For an exhaustive discussion of path planning methods we refer to \cite{gammell2021asymptotically} and \cite{elbanhawi2014sampling} for an overview on (asymptotically optimal) sampling based path planning methods.

\subsection{Planning in Dynamic environments}

Planning in dynamic environments can be roughly divided in two approaches.
First, we have reactive methods, which work with the assumption that the trajectories of the moving obstacles are unknown, whereas the second category assumes full knowledge of the obstacles' trajectories.

Reactive methods such as Execution-extended RRT~\cite{bruce2002real}, Closed-loop RRT~\cite{luders2010bounds, kuwata2008motion}, RRTX~\cite{Otte2016} or Real-time RRT*~\cite{naderi2015rt} are methods specifically developed for rapid replanning. 
Rapid replanning is necessary when previously computed paths become invalid during execution.
Risk-RRT~\cite{fulgenzi:inria-00526601} incorporates predictions about the obstacles' movement, and computes partial motion paths to keep the probability of a collision under a given threshold.
However, frequent replanning is still needed as only partial paths are returned.
Various methods exist to enable efficient replanning, i.e. to reuse as much prior work as possible from previously planned paths, or to establish coarse connectivity of the space, and only replan for dynamic obstacles (\cite{jaillet2004prm, yang2010elastic, ferguson2006replanning, zucker2007multipartite}). 

Contrary to reactive methods, the following methods assume full knowledge of obstacle trajectories, and thus do not rely on replanning.
Time-Based RRT \cite{sintov_TBRRT} expands the configuration state space by the time dimension and plans unidirectionally to a set of known goal states.
However, knowledge of the specific time for each goal configuration is assumed, and only unidirectional planning is supported.
Safe Interval Path planning \cite{phillips2011sipp} finds optimal paths with respect to shortest time by constructing a discrete search space with states defined by their configuration and a corresponding `safe interval'. 
However, a graph needs to be constructed for the entire state space, and thus it suffers the inherent problems: it is only feasible for problems with few dimensions.

In this work, we assume full knowledge of all paths of the moving obstacles, but no a priori knowledge of the arrival time, as is the case in multi-robot assembly planning tasks~\cite{Hartmann2021LongHorizonMR}.
Thus, our method does not require replanning and is able to efficiently find feasible and time-optimal paths.
Our method also enables to plan bidirectionally in unbounded time spaces, leading to a more efficient planner than other RRT-based planners in the space-time setting.

\subsection{Time-optimal Trajectory planning}

A common approach to find kinodynamically feasible paths is based on path-velocity decomposition: first find a geometrically feasible path, and then find a valid time-parametrization for this path \cite{kant1986toward}.
Extensions to this approach were presented e.g. in \cite{pham2017admissible}, which relaxes the quasi-static requirement.
However, this approach is inapplicable here, as obstacles are dynamic and the time optimization on a fixed path might render it infeasible.

Other approaches to planning include optimization approaches (e.g. STOMP \cite{kalakrishnan2011stomp}, or sequential convex optimization \cite{schulman2014motion}), or extending the configuration space with velocity coordinates \cite{webb2013kinodynamic}.
Optimization based approaches work well to incorporate complex constraints, but suffer from the well known non-convexity of the general planning problem.
Furthermore, optimizing for arrival time is not straightforward.
In general, these methods are not complete and therefore can not achieve global optimality.

Sampling based kinodynamic planning on the other hand, doubles the dimensionality of the state space we plan in, and thus makes planning with high DoF-robots slow or even infeasible.
Since time is not taken into account explicitly, planning with dynamic obstacles is not straightforward.

By extending the configuration space with a time component, and planning and optimizing in this space-time state space, we retain these guarantees.
Through usage of bidirectional planning, conditional sampling, and simplified rewiring, we achieve a high efficiency.

%% file: sections/3_algorithm.tex
We consider the motion planning problem in space-time with unbounded arrival time. 
Our objective is to minimize arrival time under given velocity constraints. 
By adding a time dimension to the configuration space we obtain the \textit{Space-Time} state-space $\X = \Q \times \T$, where $\Q$ is the underlying configuration state space and $\T$ is the time state space.
Note that $\X$ can be unbounded in time.
Let $\Xfree \in \X$ be the obstacle-free subset of states, $\xstart$ the start state, and $\Xgoal = \Qgoal \times \Tgoal$ the goal region.
In the following, we assume full knowledge of the obstacles' trajectories, and plan for holonomic robots with a given maximum velocity. 
We define $\vmax \in\mathbb{R}^{|\Q|}$ as a vector containing the maximum velocity for each space component.

\def\tend{t_{1}}
The goal is to compute a continuous path $p : [0,1] \to \Xfree$, such that $p(0) = \xstart$, $p(1) \in \Xgoal$, and the velocity constraints are satisfied. 
We are interested in finding not only feasible, but paths which minimize the arrival-time, $c(p)=\tend$ with $\tend$ being the time element of $p(1)=(q_1,\tend)$. 

In space-time, the distance that can be covered in a given time is constrained by $\vmax$ and it is not possible to move backwards in time. 
Thus, we define our distance function $d$ between two states, $x_1 = (q_1, t_1)$ and $x_2 = (q_2, t_2)$ as
\begin{equation}
\arraycolsep=2pt\def\arraystretch{1}
    d(x_1, x_2)\! =
    \begin{cases}
    \lambda d_\Q(q_1, q_2)\! +\! (1\!-\!\lambda)(t_2\!-\!t_1),\\
    &\phantom{\text{(Ooooo}}\llap{\text{ if }$t_1<t_2, ~v^i\leq v^i_{\text{max}}\ \forall i \in \left[1,|\Q|\right]$} \\
    \infty, &\text{else.}
    \end{cases}
\end{equation}
where $d_\Q$ is the intrinsic metric of the configuration space, $\lambda \in \left( 0,1 \right)$ weights the importance of $d_\Q$ with respect to the time-distance (but does not influence optimality), and $v^i$ is the required speed in dimension $i$, such that $q_2$ can be reached from $q_1$ in time $t_2 - t_1$. 
As $d$ is not symmetric, it is only a pseudometric. 

\subsection{Algorithm \label{ssec:algorithm}}

\begin{algorithm}[t]
\caption{ST-RRT*}\label{alg:main}
\begin{algorithmic}[1]
\Require $\X, \xstart, \Xgoal, d, \ptc, \tmax, \pgoal, P$
\State $T_{a} \gets \{\xstart\}\; T_{b} \gets \emptyset$
\State $B \gets \Call{InitializeBoundVariables}{P}$
\While{$\neg\ptc$}
    \State $B \gets \Call{UpdateGoalRegion}{B, P, \tmax} \label{alg:spacetimerrt:extendgoalregion}$
    \If{$\pgoal \geq \Call{Rnd}{0,1} \label{alg:spacetimerrt:pgoal}$}
        \State $B \gets \Call{SampleGoal}{\xstart, \Xgoal, \Treegoal, \tmax, B} \label{alg:spacetimerrt:samplegoal}$
    \EndIf
    \State $\xrand \gets \Call{SampleConditionally}{\xstart,\X, B, d} \label{alg:spacetimerrt:sampleconditionally}$
    \IfNot{$\Call{Extend}{T_a, \xrand, d} \label{alg:spacetimerrt:extend} = Trapped$}
        \State $\sib \pluseq 1$ 
        \State $\ts \pluseq 1$
        \State {$\Call{RewireTree}{T_a, \Treegoal, \xnew} \label{alg:spacetimerrt:RewireGoalTree}$}
        \If{$\Call{Connect}{T_b, \xnew, d} = Reached$}\label{alg:spacetimerrt:connect}
        \State $solution \gets \Call{UpdateSolution}{\xnew} \label{alg:spacetimerrt:updatesolution}$
        \State $\tmax \gets \Call{CostPath}{solution} \label{alg:spacetimerrt:costpath}$
        \State $\bp \gets 1$
        \State $\Call{PruneTrees}{\tmax, T_a, T_b}$ \label{alg:spacetimerrt:prune}
        \EndIf
    \EndIf
    \State $\Call{Swap}{T_a, T_b} \label{alg:spacetimerrt:swap}$
\EndWhile
\State \Return $solution$
\end{algorithmic}\label{alg:spacetimerrt}
\end{algorithm}


The algorithmic details of ST-RRT* are shown in Algorithms~1--5. 
In addition to $\X$, $\xstart$, $\Xgoal$, and $d$ it requires a planner termination condition $\ptc$, a time bound $\tmax \in (0, \infty]$, a probability to sample a new goal $\pgoal \in (0, 1]$, and several bound parameters contained in $P$ (see \cref{sssec:progressivegoal}).
The basic framework is similar to RRT-Connect \cite{kuffner2000rrt-connect}:
In each iteration a new goal is sampled with probability $\pgoal$ (Line~\ref{alg:spacetimerrt:pgoal}~\&~\ref{alg:spacetimerrt:samplegoal}). 
Then, a random state $\xrand$ is sampled (Line~\ref{alg:spacetimerrt:sampleconditionally}). 
If possible, the current tree $T_a$ is expanded by the new state $\xnew$ (i.e. the extension between $\xnear$ and $\xrand$) and a connection from $\xnew$ to the other tree $T_b$ is attempted (Line~\ref{alg:spacetimerrt:extend} \&~\ref{alg:spacetimerrt:connect}). 
In case of a successful connection, the solution is updated (Line~\ref{alg:spacetimerrt:updatesolution}).
Finally, $T_a$ and $T_b$ are swapped and the next iteration begins (Line~\ref{alg:spacetimerrt:swap}). 
Our extensions to RRT-Connect are: 
\begin{itemize}
    \item \textit{Progressive Goal Region Expansion}, which progressively enlarges the time component of the space (Line~\ref{alg:spacetimerrt:extendgoalregion}), and samples new goals for the goal tree (Line~\ref{alg:spacetimerrt:samplegoal}), 
    \item \textit{Conditional Sampling} (Line~\ref{alg:spacetimerrt:sampleconditionally}), which first samples a state from $\Q$, and then samples a corresponding valid time, with which $\xrand$ is constructed, and 
    \item \textit{Simplified Rewiring}, which improves the solution (Line~\ref{alg:spacetimerrt:RewireGoalTree}) by optimizing for minimal arrival time.
\end{itemize}
We also prune the trees (Line~\ref{alg:spacetimerrt:prune}) to remove parts which cannot improve the solution anymore. 

\subsubsection{Progressive Goal Region Expansion\label{sssec:progressivegoal}}
\begin{figure}
    \centering
    \includegraphics[width=.9\linewidth, height=5cm]{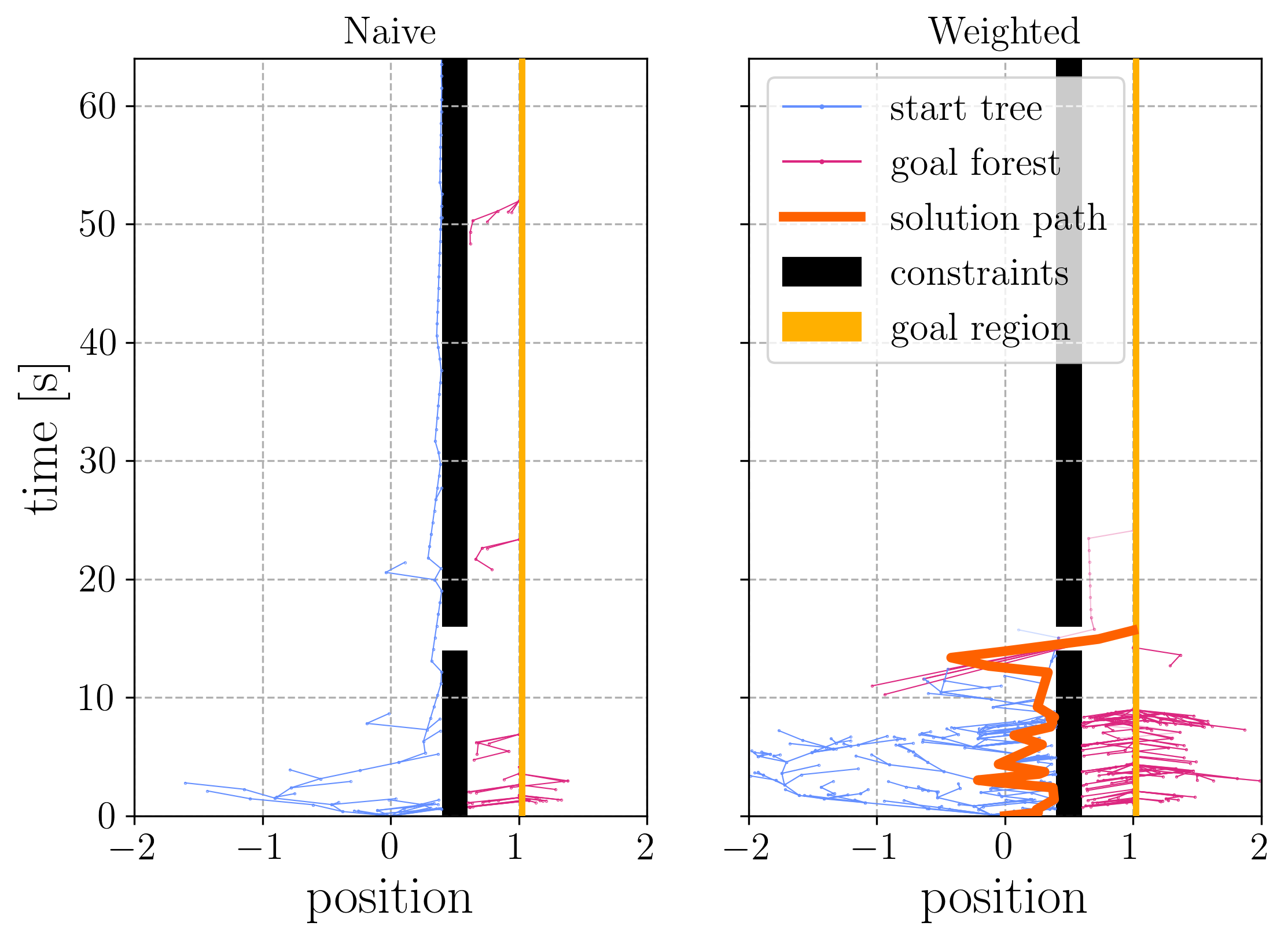}
    \caption{Illustration of the search trees after the same computation time with naive and weighted sampling strategy with similar numbers of samples (for naive sampling, not all samples are visible, and the time bound was increased beyond the shown range).
    }
    \label{fig:goals}
\end{figure}
If the time-space $\T$ is unbounded it is difficult to generate samples distributed throughout the whole space.
However, when imposing an arbitrary time-bound, the problem might become infeasible \cite{gammell2015batch, gammell_phd17}.
Therefore, we expand the sampled goal region progressively whenever a new batch of samples is added.
To do that, we introduce several parameters contained in the bound struct $B$:
$\tr$ determines the time bound for goal sampling and $\bs$ determines after how many generated samples the expansion takes place.
When a batch is full, $\tr$ is increased by $\rf$ and $\bs$ is increased accordingly.

With an increasing time-bound, the sample density is higher at the lower time values due to the previously generated samples.
\Cref{fig:goals} shows how naive sampling may lead to cases where it becomes increasingly unlikely to find any solution. 
Thus we use weighted sampling, where the old and newly expanded region are explicitly sampled with probability $\bp$ and $1 - \bp$, respectively, to ensure a uniform distribution over the total space.\looseness=-1

Precisely, the \textit{Progressive Goal Region Expansion} works as follows: The parameters $\rf$, $\ibs$, and $\sr$ are user-specified. 
All variables of $B$ are initialized at the start (\cref{alg:init}) and updated during execution.
While $\tr$ is used when the current goal region is sampled, $\ntr$ is used to sample the newly expanded one.
After the first expansion, $\ntr$ is always higher than $\tr$ by a factor equal to $\rf$ (Alg.~\ref{alg:extend}, Line~\ref{alg:extend:timerange}~\&~\ref{alg:extend:newtimerange}).
The minimum amount of the new batch size is given by $(\rf - 1) \cdot \ts$.
That is, when all samples of the new batch are placed in the new region, the overall distribution would be uniform over the time-space. 
To ensure that the old region is also sampled, $\bs$ is further increased by $\sr \in (0,1)$ (Line~\ref{alg:extend:batchsize}).
The probability to sample the old batch $\bp$ is calculated in dependence of $\rf$ and $\sr$ (Line~\ref{alg:extend:prob}).
Due to the exponential growth of the batch size, the choice of the configuration parameters is important for performance.

To sample a goal state, its space component $q$ is sampled first (Alg~\ref{alg:samplegoal}, Line~\ref{alg:samplegoal:samplespacecomponent}). 
The lower and upper bounds for the time, $\tlb$ and $\tub$, are calculated in dependence on whether the time is explicitly bounded (Line~\ref{alg:samplegoal:if}), the current region is sampled (Line~\ref{alg:samplegoal:elseif}), or the newly expanded one is sampled (Line~\ref{alg:samplegoal:else}).
The sampling of nongoal-states is subject to the sampled goal states and therefore implicitly bounded by the time value of the sampled goal states (\cref{sssec:conditionalsampling}). 

\begin{algorithm}
\caption{InitializeBoundVariables}\label{alg:init}
\begin{algorithmic}[1]
\Require $P$
\State $\tr \gets \rf$
\State $\ntr \gets \rf$
\State $\bs \gets \ibs$
\State $\sib \gets 0\; \ts \gets 0$
\State $\bp \gets 1$
\State $\g \gets \emptyset\; \ng \gets \emptyset$
\State \Return $B$

\end{algorithmic}
\end{algorithm}

\begin{algorithm}
\caption{UpdateGoalRegion}\label{alg:extend}
\begin{algorithmic}[1]
\Require $B, P, \tmax$
\If{$\tmax = \infty$ \textbf{and} $\sib = \bs \label{alg:spacetimerrt:b}$}
    \State $\tr \gets \ntr \label{alg:extend:timerange}$
    \State $\ntr \asteq \rf \label{alg:extend:newtimerange}$
    \State $\bs \gets \frac{(\rf - 1)\ts}{\sr} \label{alg:extend:batchsize}$
    \State $\bp \gets \frac{1 - \sr}{\rf} \label{alg:extend:prob}$
    \State $\g \gets \g \cup \ng$
    \State $\ng \gets \emptyset\; \sib \gets 0$
\EndIf
\State \Return $B$

\end{algorithmic}
\end{algorithm}

\begin{algorithm}
\caption{SampleGoal}\label{alg:samplegoal}
\begin{algorithmic}[1]
\Require $\xstart, \Xgoal, \Treegoal, \tmax, B$
\State $q \gets \Call{Sampleuniform}{\Qgoal} \label{alg:samplegoal:samplespacecomponent}$
\State $\tmin \gets \Call{\funcLowerBoundArrivalTime}{\qstart, q} \label{alg:samplegoal:timedistance}$
\State $\textsc{SampleOldBatch}\!\gets\! \Call{Rnd}{0, 1}\! \leq\! \bp$
\If{$\tmax \neq \infty \label{alg:samplegoal:if}$}
    \State $\tlb \gets \tmin\; \tub \gets \tmax$
\ElsIf{$\textsc{SampleOldBatch} \label{alg:samplegoal:elseif}$}
    \State $\tlb \gets \tmin\; \tub \gets \tmin \cdot \tr \label{alg:samplegoal:r1}$
\Else \label{alg:samplegoal:else} 
    \State $\tlb \gets \tmin \cdot \tr$
    \State $\tub \gets \tmin \cdot \ntr$
\EndIf
\If{$\tub > \tlb$}
    \State $t \gets \Call{SampleUniform}{\tlb, \tub} \label{alg:samplegoal:sampleuniform}$
    \State $\Treegoal \gets \Treegoal \cup \{(q, t)\}$
    \If{$\textsc{SampleOldBatch}$}
        \State $\g \gets \g \cup \{(q, t)\}$
    \Else
        \State $\ng \gets \ng \cup \{(q, t)\}$
    \EndIf
    \State \Return $B\label{alg:samplegoal:return}$
\EndIf

\end{algorithmic}
\end{algorithm}

\subsubsection{Conditional Sampling\label{sssec:conditionalsampling}}
\begin{figure}[t]
    \centering
    \includegraphics[width=0.9\linewidth]{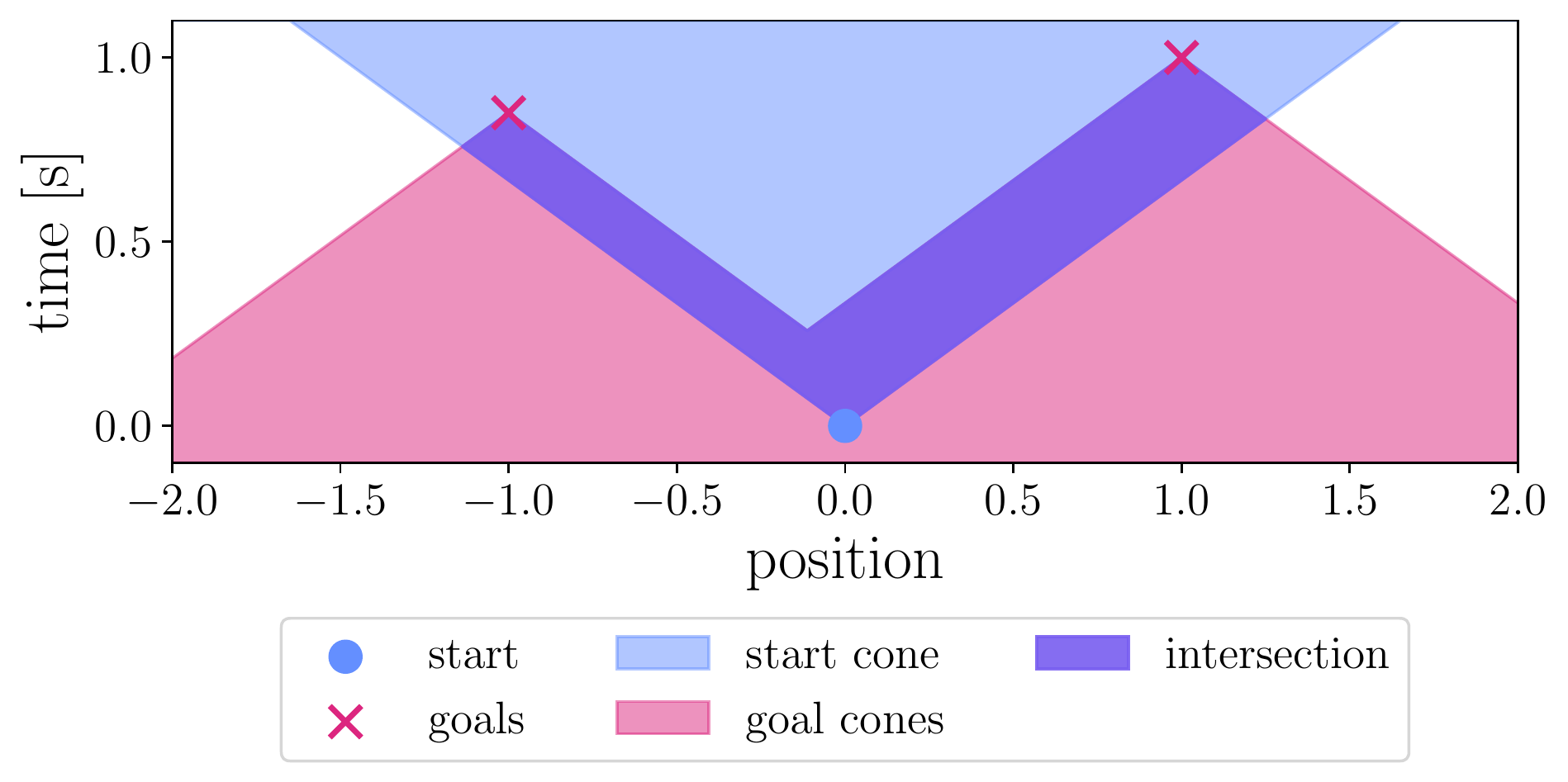}
    \caption{The start and goal cones contain all states that can be reached from the start or can reach the goal respectively. The intersection contains all states that can be part of a solution.}
    \label{fig:cones}
\end{figure}
Any state that can be part of a solution path must have a finite distance $d$ to the start and at least one goal state. 
Due to velocity-constraints, only states in the intersection of the start and goal cones (see \cref{fig:cones} for an illustration) meet this requirement.
Thus, similar to Informed RRT* \cite{gammell2014informed}, we only sample the region that can produce solutions.
Ideally, one would sample directly from the union of intersections of start and goal velocity-cones.

However, as the explicit computation of the intersection is not possible for multiple goal states, we use \textit{Conditional Sampling}: We first uniformly sample a configuration $q$ (Alg~\ref{alg:samplecondiitonally}, Line~\ref{alg:sampleconditionally:samplespacecomponent}).
Using $q$, we then sample a feasible time from the range of possible times conditioned upon $q$. 
The range of possible times is dependent on $\xstart$ and the previously sampled goal states. 
To sample more uniformly, we use two goal sets:
$\g$ for the current goal states and $\ng$ for the goal states in the newly expanded region.
The time bounds $\tlb$, $\tub$ are obtained by the minimal arrival time from the start configuration $\qstart$ until $q$ (Line~\ref{alg:sampleconditionally:tmin}) and the maximum valid time given by:
\begin{align}
    \textsc{MaxValidTime}(q_{\text{rnd}},G) &=\notag\\ \max_{(q_g, t_g) \in G}& \left(  t_g - \min_i\frac{d_\Q(q^i_\text{rnd}, q^i_g)}{v^i_{\text{max}}}\right)
    \label{eqn:max}
\end{align}
The specific calculation of $\tlb$, $\tub$ is dependent on whether the current (Line~\ref{alg:sampleconditionally:if}) or the new region is sampled (Line~\ref{alg:sampleconditionally:else}).



\begin{algorithm}
\caption{SampleConditionally}\label{alg:samplecondiitonally}
\begin{algorithmic}[1]
\Require $\xstart,\X, B$
\Repeat
\State $q \gets \Call{SampleUniform}{\Q} \label{alg:sampleconditionally:samplespacecomponent}$
\State $\tmin \gets \tstart + \Call{\funcLowerBoundArrivalTime}{\qstart, q} \label{alg:sampleconditionally:tmin}$
\If{$\Call{Random}{0,1} < \bp \label{alg:sampleconditionally:if}$}
    \State $\tlb \gets \tmin$
    \State $\tub \gets \textsc{MaxValidTime}(q,\g)
    \label{alg:sampleconditionally:rold}$
    \Comment{eq~(\ref{eqn:max})}
\Else \label{alg:sampleconditionally:else}
     \State $\tmin^* \gets \textsc{MaxValidTime}(q,\g) \label{alg:sampleconditionally:lnew}$
    \State $\tlb \gets \Call{Max}{\tmin, \tmin^*}$
    \State $\tub\gets \textsc{MaxValidTime}(q,\ng) \label{alg:sampleconditionally:rnew}$
\EndIf
\Until{$\tlb < \tub$}
\State $t \gets \Call{SampleUniform}{\tlb, \tub} \label{alg::sampleconditionally:sampleuniform}$
\State \Return $(q,t)$

\end{algorithmic}
\end{algorithm}

\subsubsection{Simplified Rewiring\label{sssec:rewiring}}
To compute time-optimal solutions ST-RRT* uses similar methods as RRT* and preserves its property of asymptotic optimality. 
Equal to RRT*, ST-RRT* tries to rewire a set of states near to the newly added state, $\xnew$, after tree expansion. 
Contrary to RRT*, rewiring is only performed in the goal trees.
This is due to the fact that rewiring nodes in the start tree can never lead to a better arrival time in the path.
Rewiring states in the start tree can not change their arrival time, whereas in the goal trees a node can be rewired to a root node with a smaller time value. 
One more deviation is the check of which nodes should be rewired.
For all nodes in the goal trees simply the time value of their respective root node has to be considered.

%

\subsection{Proof Sketches}

To prove probabilistic completeness in space-time, we distinguish between two cases. 
In case of bounded time, planning with a quasi-metric reverts to kinodynamic planning, where we refer to results from~\cite{kleinbort2018probabilistic} and~\cite{janson2015fast} for completeness proofs.

The second case is unbounded time:
If a solution exists, there needs to be a feasible goal region at a finite time. 
Since we iteratively increase the upper bound, we will, eventually, have increased the goal region to include the feasible goal region. 
Due to the use of uniform sampling of the time range, there will be positive probability that the feasible goal region will be sampled.
Since \textit{conditional sampling} always gives a positive probability of sampling any open set, this makes ST-RRT* retain probabilistic completeness~\cite{janson2018deterministic}.

Apart from probabilistic completeness, ST-RRT* is also asymptotically optimal with respect to arrival time.  
Since ST-RRT* is modelled after RRT-Connect, it can be made asymptotically optimal by tree rewiring~\cite{karaman2011sampling, Salzman2016}. Inside the rewiring step, we connect newly added states to the nearest goal tree which minimizes arrival time. This ensures asymptotic optimality with respect to final arrival time.

%% file: sections/4_evaluation.tex

We compared ST-RRT* to other planners on 4 different scenarios using the benchmarking capabilities of OMPL~\cite{moll2015benchmarking-motion-planning-algorithms}.
All evaluations were performed over $100$ runs with different pseudorandom seeds of $30\text{s}$ each (if not stated otherwise).
ST-RRT* is compared to RRT-Connect\footnote{The metric had to be changed to be symmetric for distance calculation, but remained as stated for motion validation (this change did not help in the other planners).} and RRT* in space-time using their OMPL implementations in default configuration.
Since RRT* and RRT-Connnect algorithms can not operate on unbounded time, three different time bounds are measured.
The lowest time bound was determined according to the best solutions of ST-RRT* and set to a higher value to ensure feasibility.
Without knowing a solution this is generally not possible.
For planning through Space-Time, most of the planners in OMPL~\cite{sucan2012the-open-motion-planning-library} do not work either due to only working with metric spaces, only working with euclidean spaces, not supporting asymmetric distance function (e.g. due to using undirected graph structures), or were never able to find solutions in the specified runtime.

\subsection{Scenarios}\label{ssec:scenarios}
\begin{figure}
\centering
    \begin{subfigure}[t]{.24\textwidth}
        \centering
        \includegraphics[width=.9\linewidth]{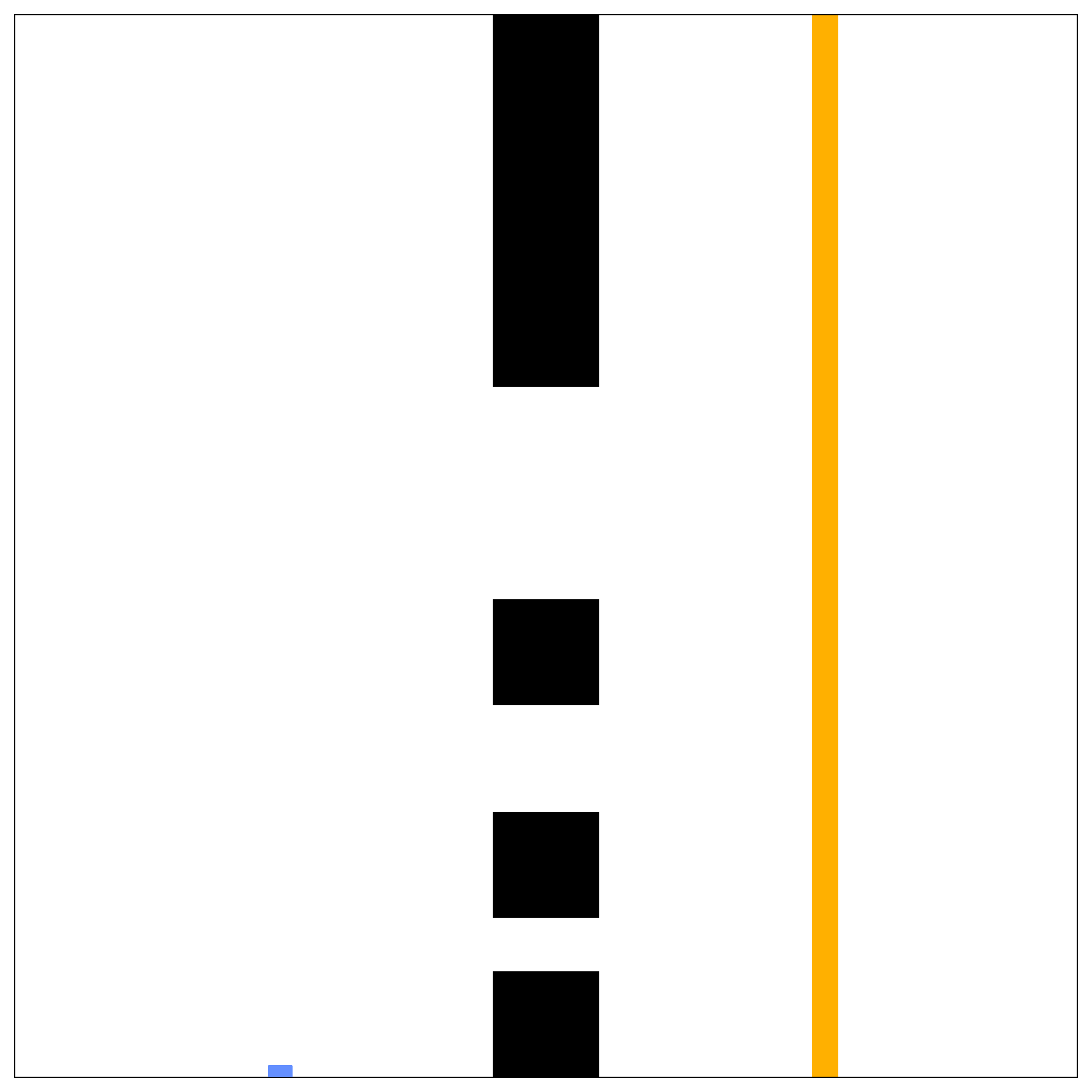}
        \caption{\label{fig:scenarios_narrow}Narrow passage in time ($\mathbb{R}^{1+1}$).}
    \end{subfigure}\hfill
    \begin{subfigure}[t]{.24\textwidth}
        \centering
        \includegraphics[width=.9\linewidth]{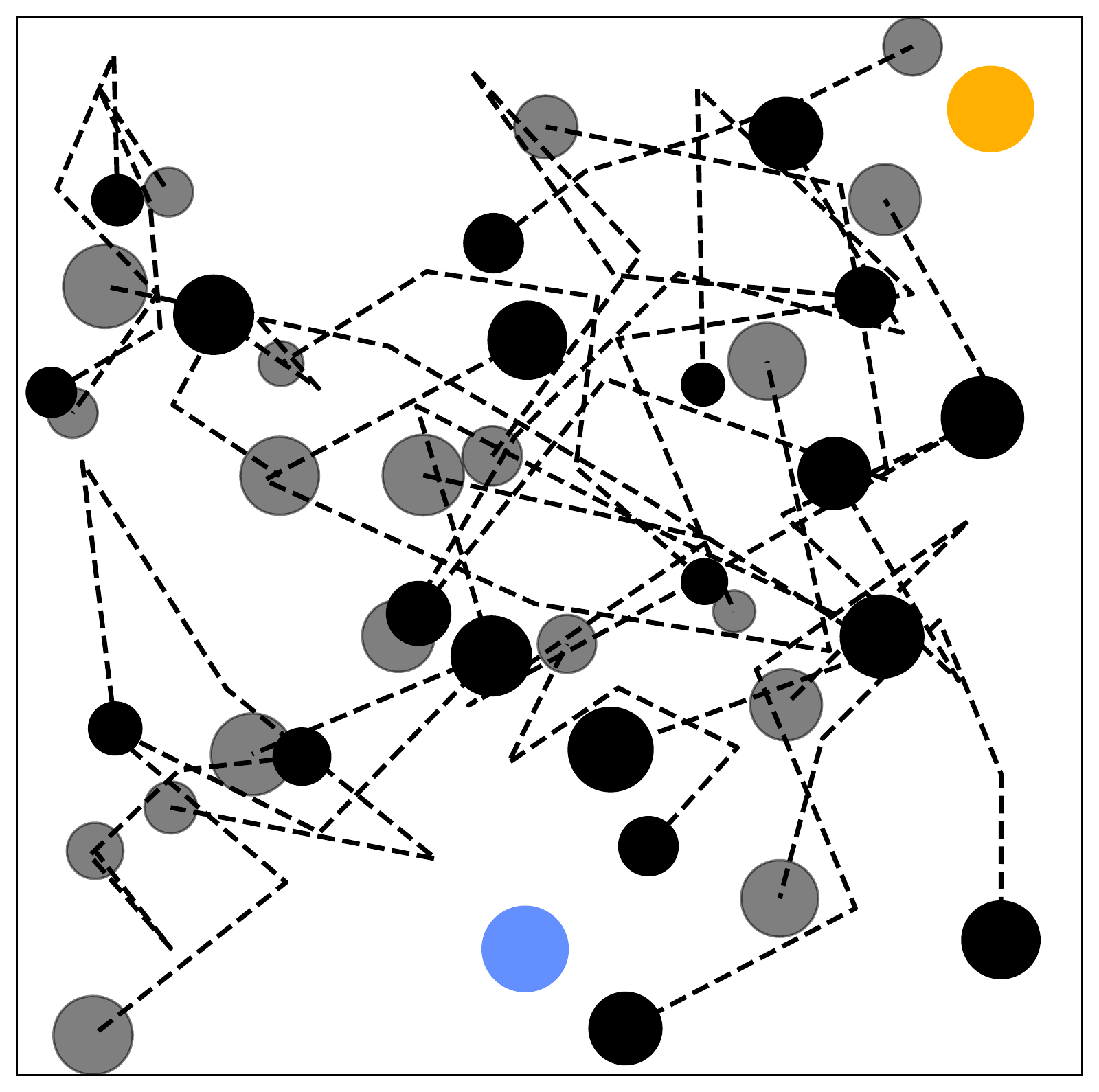}
        \caption{\label{fig:scenarios_rnd}Rnd. moving obstacles ($\mathbb{R}^{2+1}$). Obstacle start in black, end position in grey.} 
    \end{subfigure}\hfill
    
    \begin{subfigure}[t]{.24\textwidth}
        \centering
        \includegraphics[width=.97\linewidth]{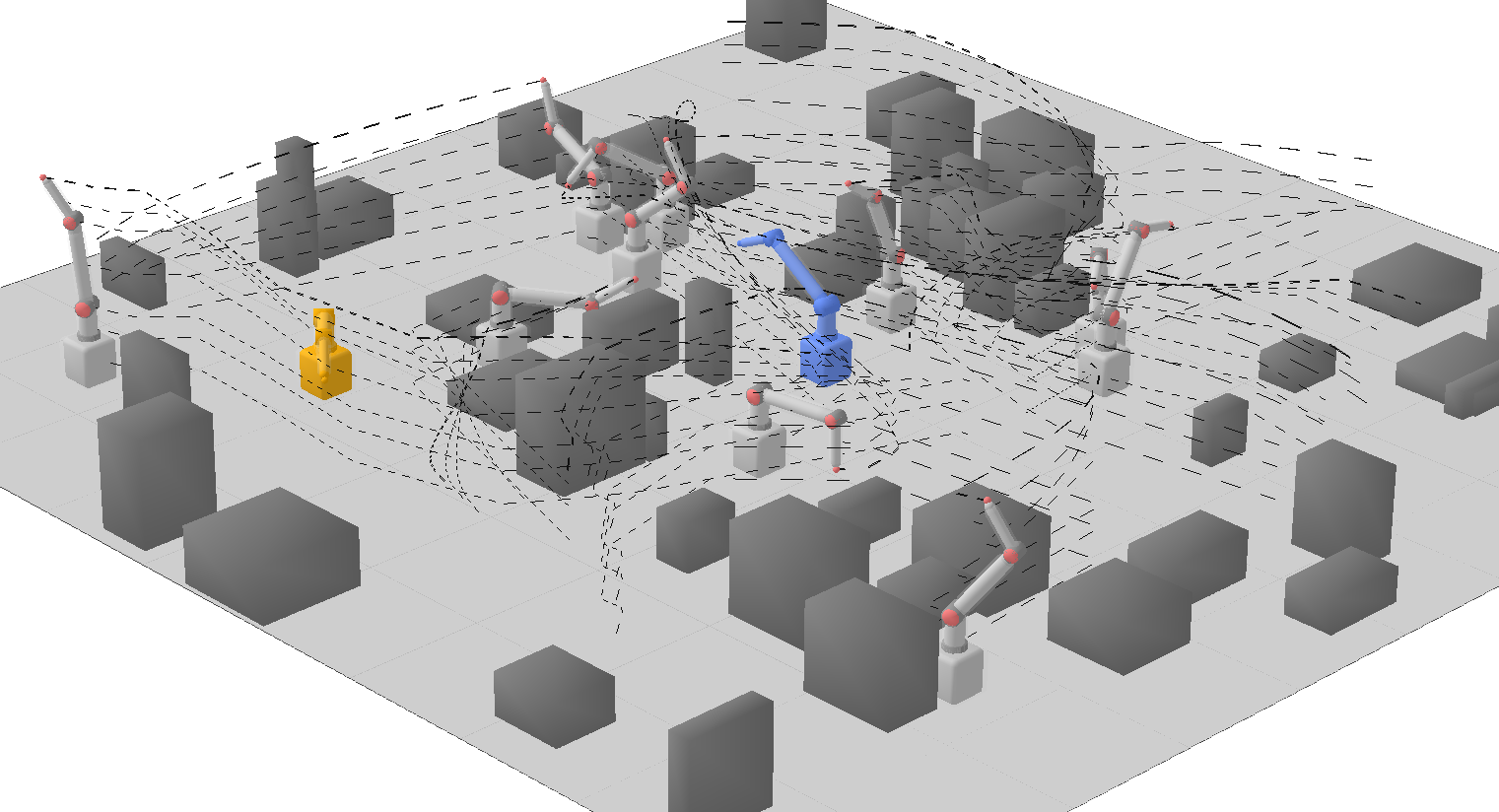} 
        \caption{\label{fig:scenarios_mobile}Mobile robots.}
    \end{subfigure}\hfill
    \begin{subfigure}[t]{.24\textwidth}
        \centering
        \includegraphics[width=.97\linewidth]{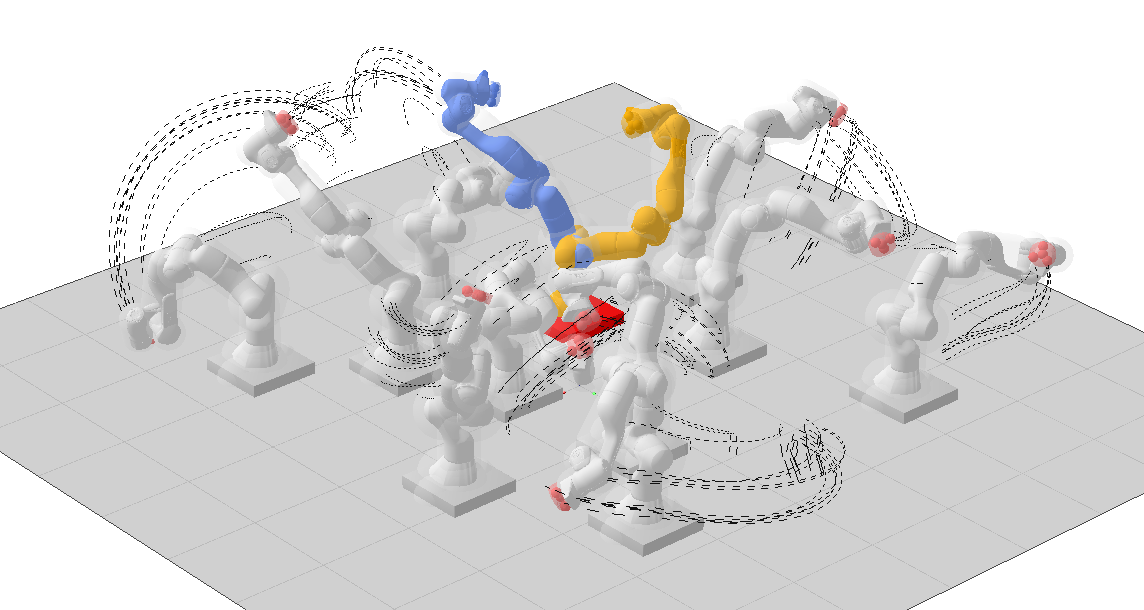}
        \caption{\label{fig:scenarios_pandas}Robotic arms.}
    \end{subfigure}
    \caption{Illustrations of the scenarios: Starts are shown in blue, goals and goal regions in yellow, and obstacles in black. The dashed lines are the paths of the moving obstacles.}
    \label{fig:scenarios}
\end{figure}
We evaluate the method on the following scenarios\footnote{Videos of the scenarios, and the paths are in the supplementary material.}:

\begin{enumerate}[(i)]
    \item \textit{Narrow passage}: A point has to move from start configuration $q_0$ to goal configuration $q_F$ in an environment where the configuration space is split into two parts by an obstacle up to a certain point in time except for three narrow periods of time (\cref{fig:scenarios_narrow}).
    \item \textit{Cluttered space}: A (hyper-)sphere has to move from $q_0$ to $q_F$ in an environment with randomly moving obstacles (\cref{fig:scenarios_rnd}).
    \item  \textit{Sequential mobile robot planning}: A robot with a mobile base and a robot arm on top ($\mathbb{R}^8$) has to move from $q_0$ to $q_F$ in an environment with randomly distributed obstacles, and other moving mobile robots that move on a fixed trajectory (\cref{fig:scenarios_mobile}). This is a common subproblem in prioritized multi robot planning~\cite{Orthey2020IJRR}.
    \item  \textit{Sequential robot arm planning}: A robotic arm ($\mathbb{R}^7$) has to move from configuration $q_0$ to $q_F$ in an environment with previously planned panda robotic arms (\cref{fig:scenarios_pandas}). Such a scenario may arise in e.g. simultaneous bin-picking with multiple robots.
\end{enumerate}
We show the narrow passage problem in $\mathbb{R}^{1+1}$ and $\mathbb{R}^{8+1}$ and the cluttered env. in $\mathbb{R}^{2+1}$ and $\mathbb{R}^{8+1}$.
For the robotic settings, we test the planners in the 6$^\text{th}$ and the 11$^\text{th}$ agent (i.e. the previous 5, and 10 agents, respectively, already have a trajectory).

\input{sections/table_results}

\subsection{Experimental Results}

We analyze the results of both the abstract experiments (\cref{fig:exp_narrow_1} - \cref{fig:exp_rnd_8}), and the simulated robot experiments (\cref{fig:exp_mobile_6} - \cref{fig:exp_ra_11}).
We compare the success rates and the cost-convergence plots of the different algorithms.


\subsubsection{Initial solution time}
In almost all cases the median initial solution time of ST-RRT* is lower than for both RRT-Connect and RRT*, even with the tightest time-bound.
This can be attributed to the conditional sampling, which helps avoid exploring areas that are clearly not reachable. 

\subsubsection{Success Rate} A low time bound helps to more quickly find solutions for RRT-Connect and RRT*; however, it can lead to the inability to find solutions at all.
This is especially problematic for RRT-Connect which stops sampling goal states at some point, leading to RRT-Connect sometimes not reaching 100\% success rate even though the time bound is specified such that a solution would be attainable.

\subsubsection{Cost} ST-RRT* converges to the best found solution more quickly than RRT*. 
Additionally, while the initial cost of the solution of ST-RRT* is sometimes higher than RRT-Connect's solution, the final solution cost of ST-RRT* is in all cases lower or equal than for the other methods.

Summarizing the results, a special treatment of the time-space is clearly necessary in a planner to achieve good performance in the motion planning process and ST-RRT* outperforms the other planners on the tested problems.

%% file: sections/table_results.tex
\begin{figure*}[!h]
\centering
    \begin{subfigure}[t]{.25\textwidth}
        \centering
        \includegraphics[width=.92\linewidth]{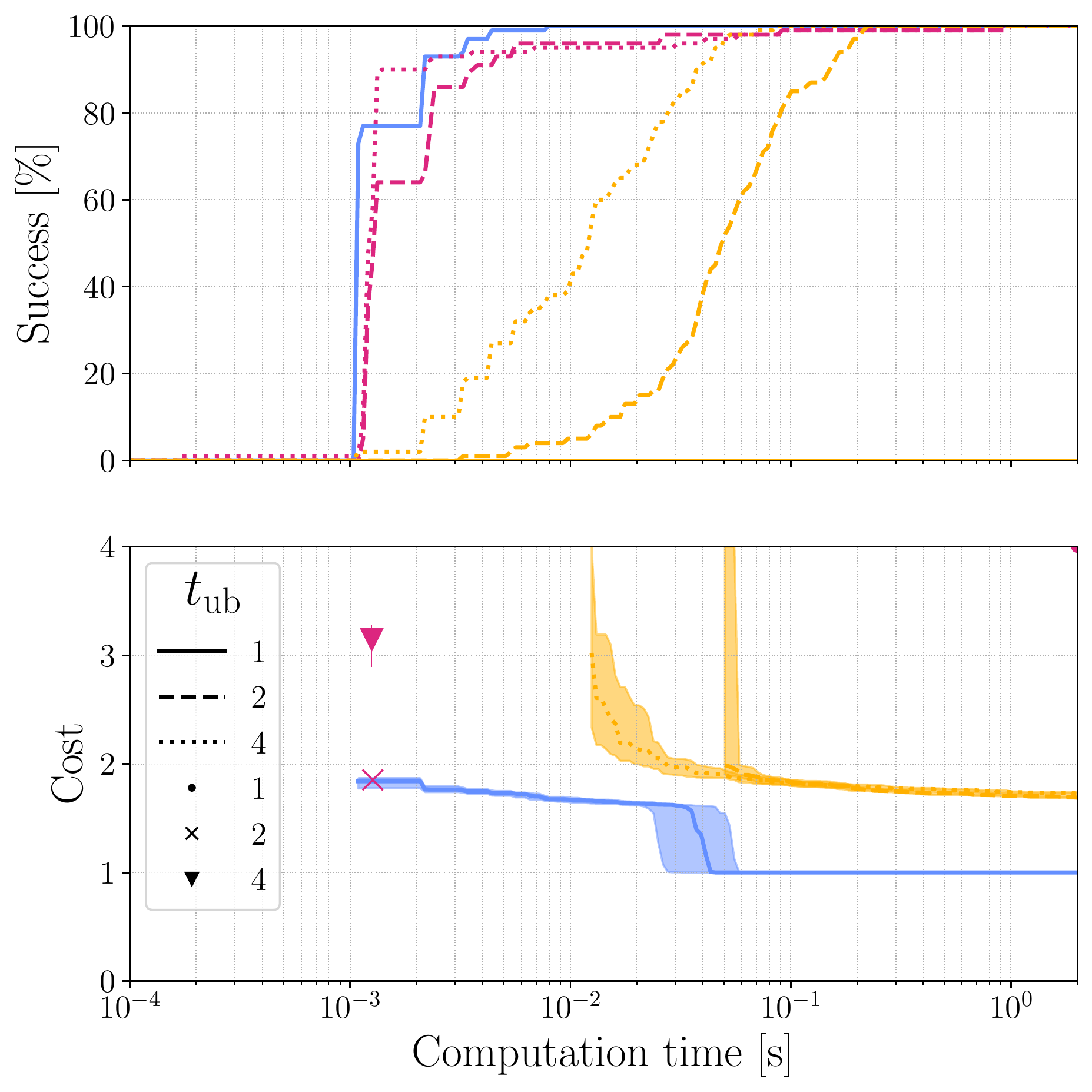} 
        \caption{\label{fig:exp_narrow_1}Narrow passage in time: $\mathbb{R}^{1+1}$}
    \end{subfigure}%
    \begin{subfigure}[t]{.25\textwidth}
        \centering
        \includegraphics[width=.92\linewidth]{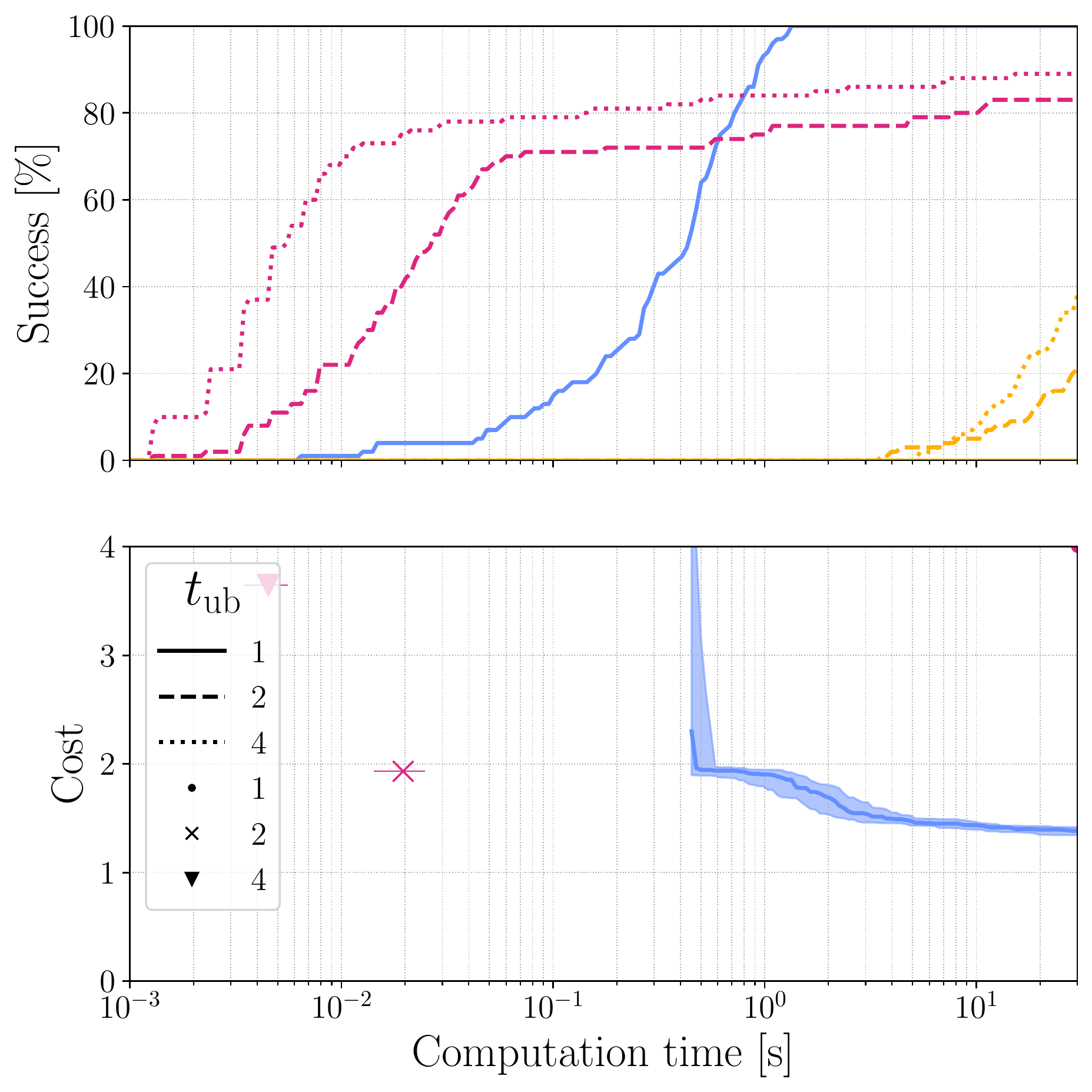}
        \caption{\label{fig:exp_narrow_8}Narrow passage in time: $\mathbb{R}^{8+1}$}
    \end{subfigure}%
    \begin{subfigure}[t]{.25\textwidth}
        \centering
        \includegraphics[width=.92\linewidth]{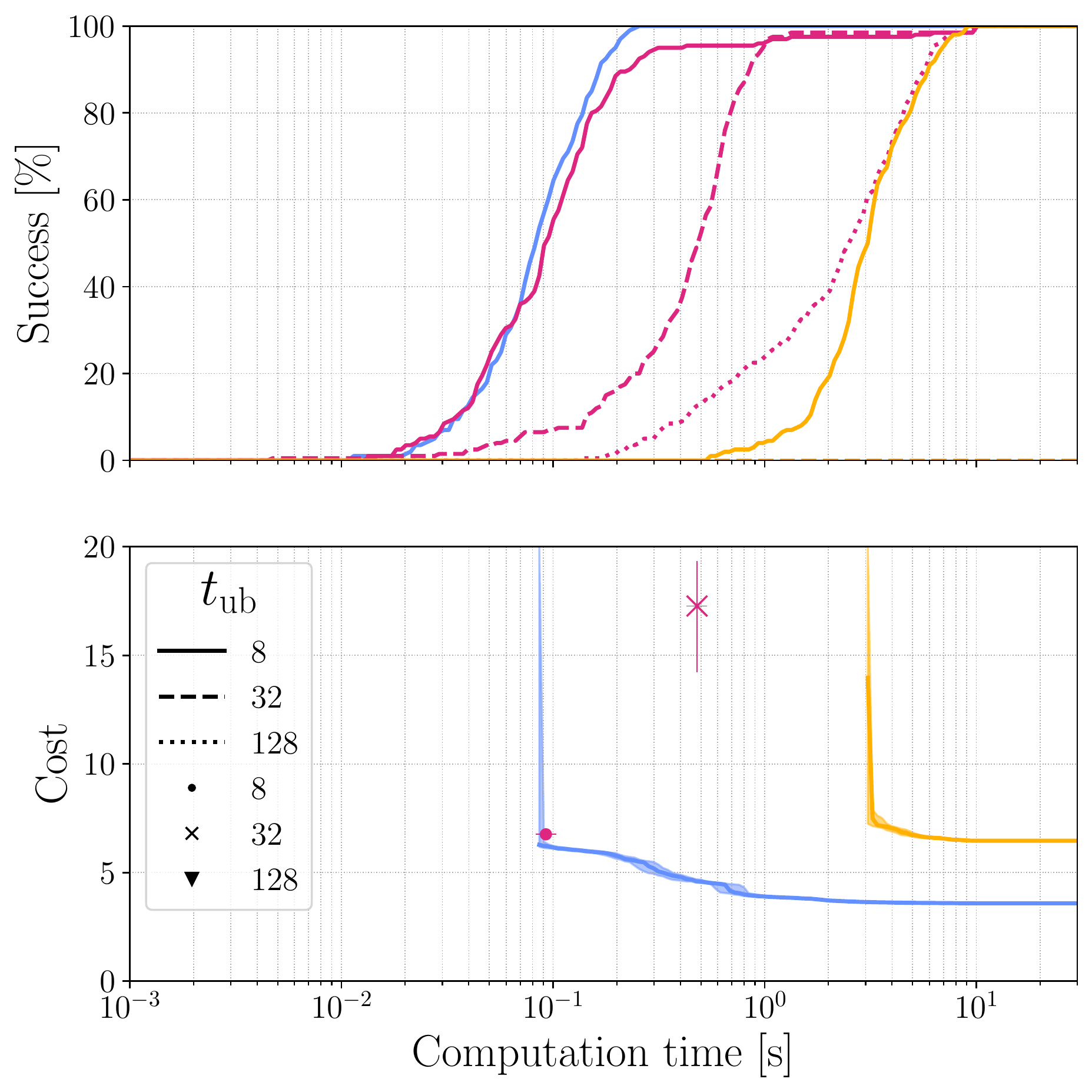} 
        \caption{\label{fig:exp_rnd_2}Rnd. moving obstacles: $\mathbb{R}^{2+1}$}
    \end{subfigure}%
    \begin{subfigure}[t]{.25\textwidth}
        \centering
        \includegraphics[width=.92\linewidth]{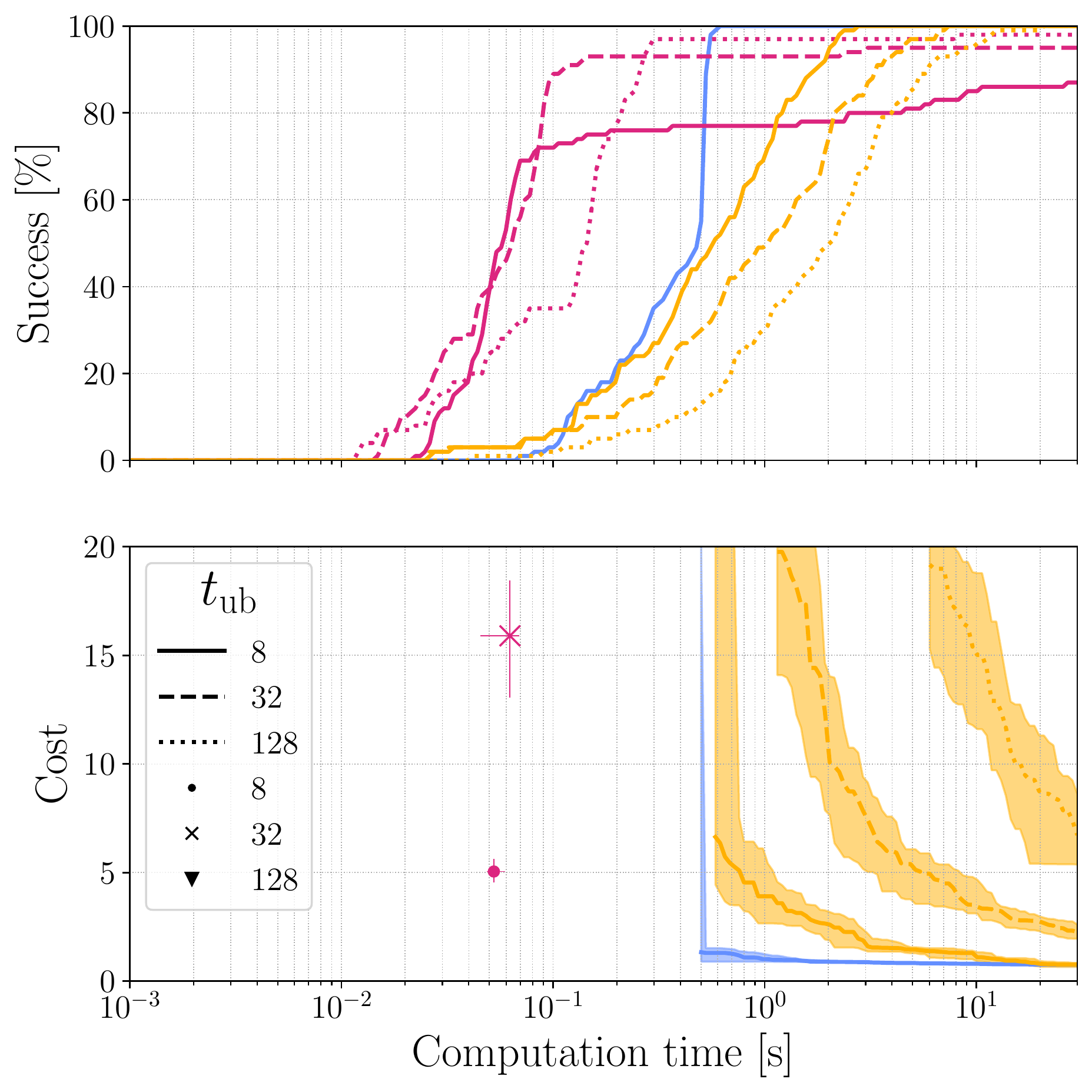} 
        \caption{\label{fig:exp_rnd_8}Rnd. moving obstacles: $\mathbb{R}^{8+1}$}
    \end{subfigure}
    
    \centering
    \begin{subfigure}[t]{.25\textwidth}
        \centering
        \includegraphics[width=.92\linewidth]{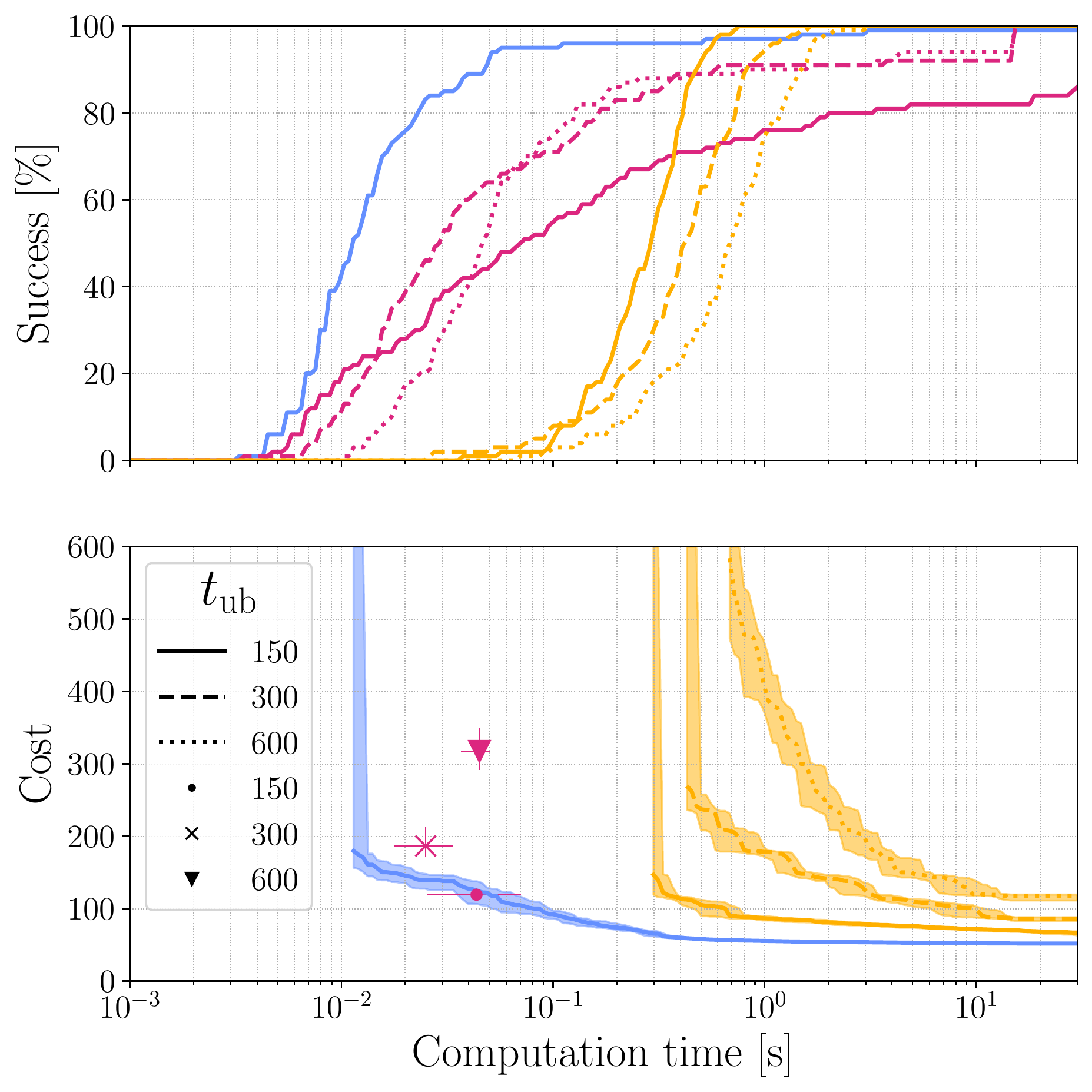} 
        \caption{\label{fig:exp_mobile_6}Mobile robots: 6$^{\text{th}}$ agent}
    \end{subfigure}%
        \begin{subfigure}[t]{.25\textwidth}
        \centering
        \includegraphics[width=.92\linewidth]{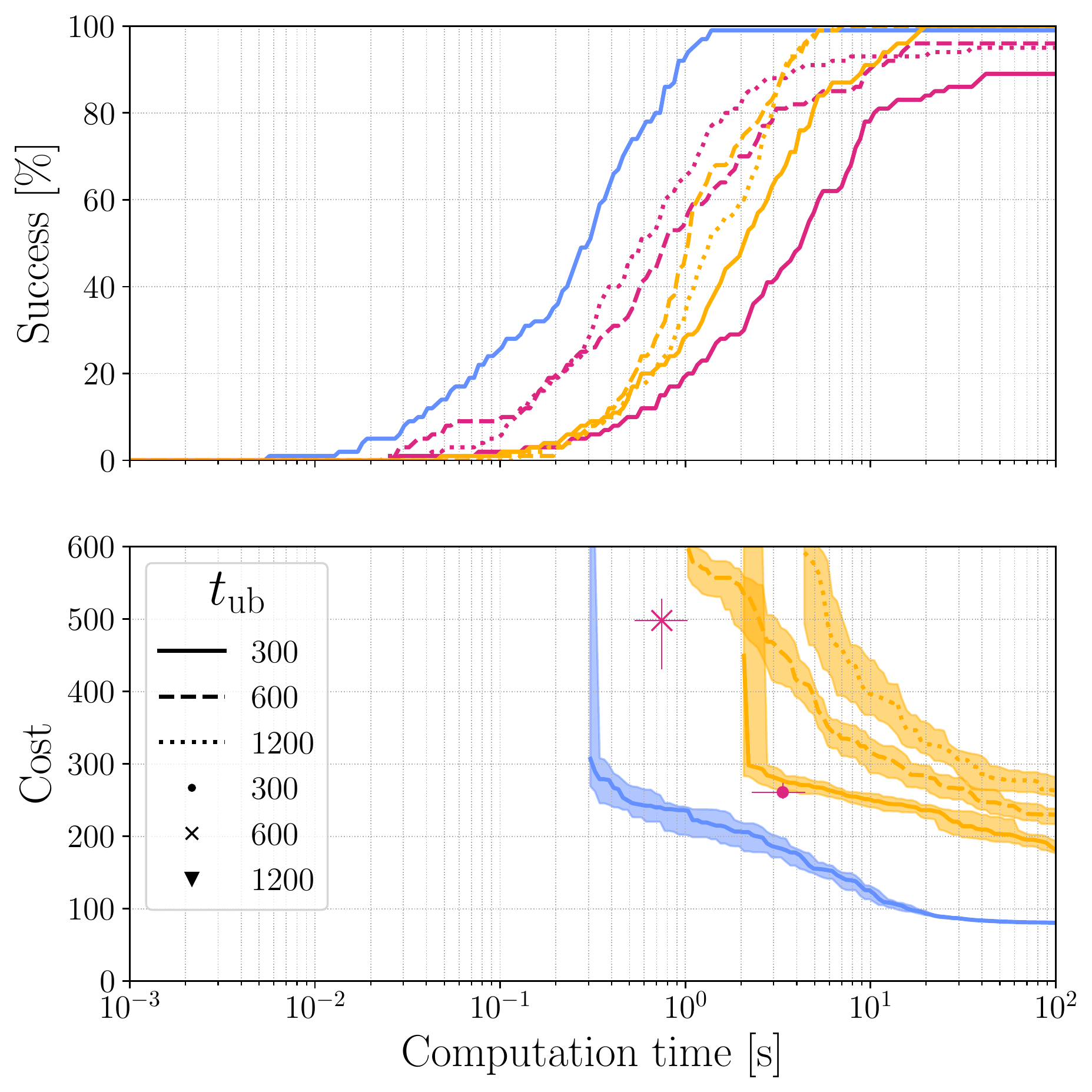} 
        \caption{\label{fig:exp_mobile_11}Mobile robots: 11$^{\text{th}}$ agent: $100\text{s}$}
    \end{subfigure}%
    \begin{subfigure}[t]{.25\textwidth}
        \centering
        \includegraphics[width=.92\linewidth]{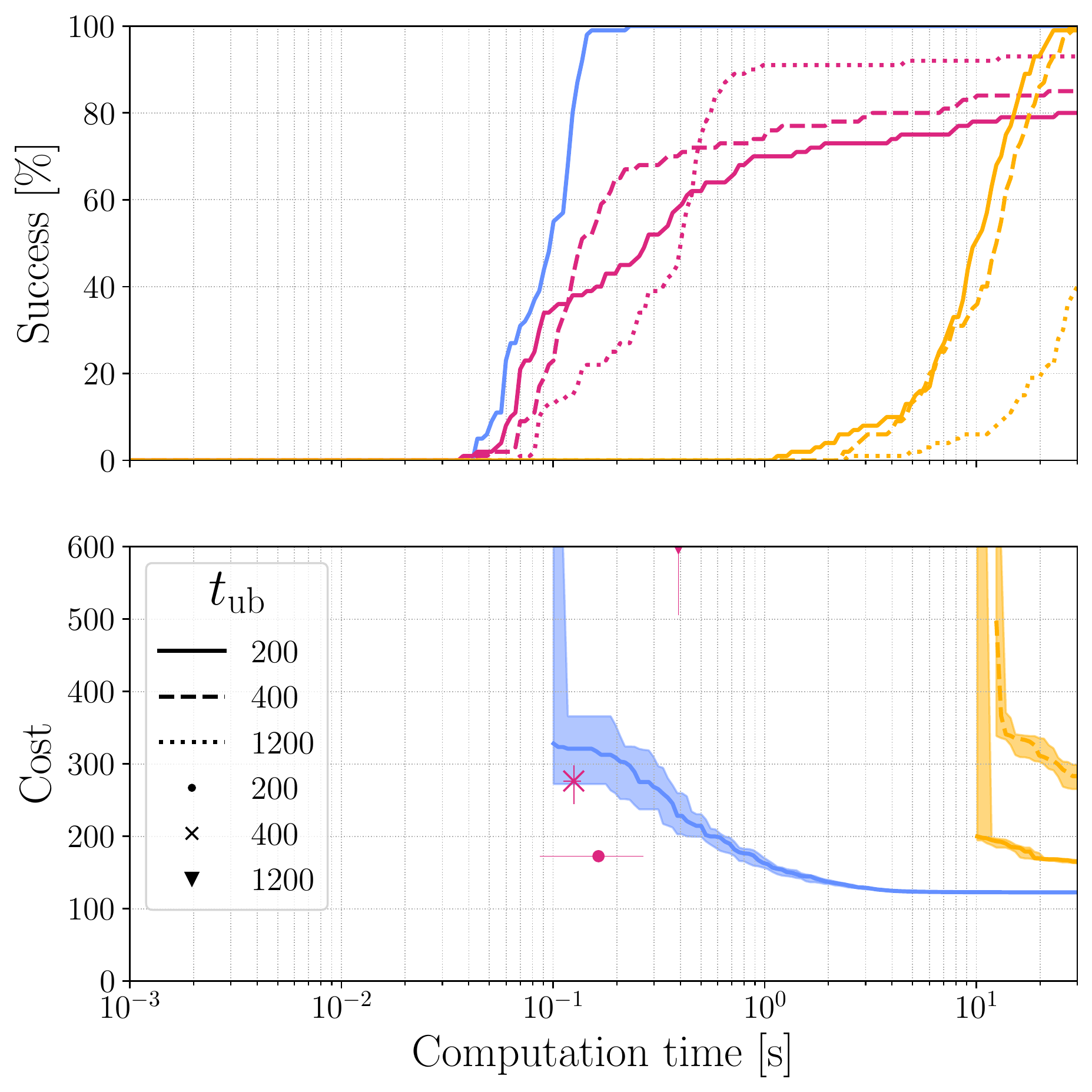}       
        \caption{\label{fig:exp_ra_6}Robot arms: 6$^{\text{th}}$ arm}
    \end{subfigure}
    \begin{subfigure}[t]{.25\textwidth}
        \centering
        \includegraphics[width=.92\linewidth]{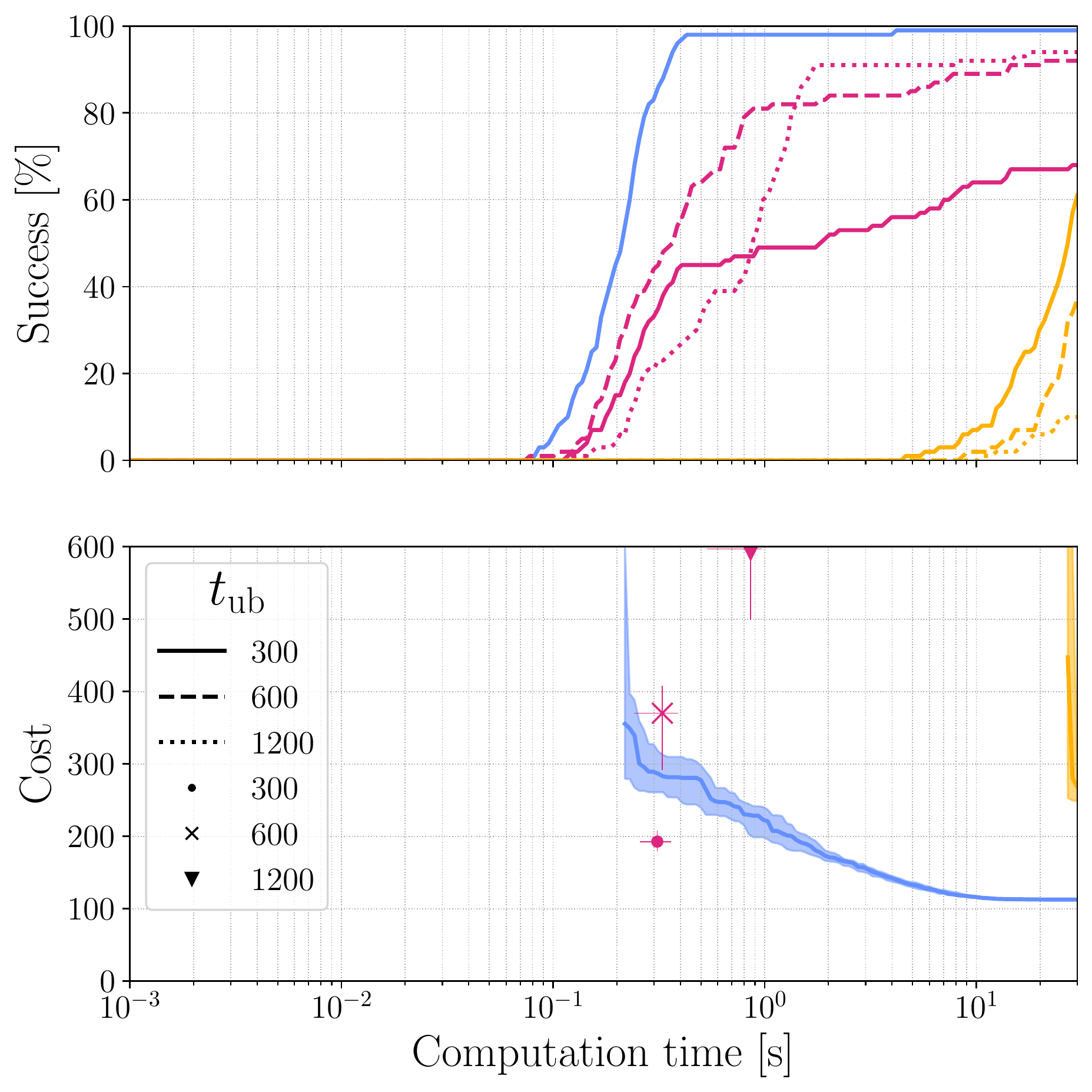}       
        \caption{\label{fig:exp_ra_11}Robot arms: 11$^{\text{th}}$ arm}
    \end{subfigure}
    \caption{Success rates and cost plots for the experiments (\cref{ssec:scenarios}) for \sqbox{cblue} ST-RRT*, \sqbox{cred} RRT-Connect, and \sqbox{cyellow} RRT* over 100 runs. 
    RRT-Connect and RRT* were run with 3 different upper bounds, $t_\text{ub}$ for the time (indicated in the figure), since they can not operate in unbounded time-spaces.
    The thick line is the median, and the shaded area in the cost plot shows the 95\% nonparametric confidence interval. 
    Cost for RRT-Connect is shown as the median with error bars for the 95\% nonparametric confidence interval.
    Unsuccessful runs are treated as infinite cost. 
    The upper time limits for RRT* and RRT-Connect are listed in the figures. 
    Planners that are not shown were not able to find any solution in the given time. 
    }
    \label{fig:exp}
\end{figure*}

%% file: sections/5_conclusion.tex
We proposed ST-RRT*, a planning algorithm that is able to efficiently deal with unbounded time spaces and optimizes for arrival time in an environment with moving obstacles on known trajectories.
We guarantee probabilistic completeness and asymptotic optimality by introducing progressive expansion of the goal space and generate new samples accordingly.
Our algorithm efficiently deals with many goals and converges to the optimal path quickly by making use of conditional sampling and shrinking the goal spaces.

The current implementation of ST-RRT* still has two limitations: the batch size and the expansion factor must be chosen in the beginning with a crude estimate of when the goal can be reached.
In practice this is not a large limitation since real settings usually impose some upper limit on the acceptable maximum time to reach a goal state.
Additionally, acceleration and more complex kinodynamic constraints (e.g. torque limits) are not taken into account. 
While this does not pose a problem in our applications, it would not be applicable to robots which have to be in quasi-static equilibrium.

We experimentally demonstrated that ST-RRT* scales well to high dimensions on both abstract and simulated robotic experiments.
Our algorithm outperforms state of the art algorithms on both initial solution time and convergence to the optimal solution.
An initial version of ST-RRT* was used in work on large-scale multi-robot coordination~\cite{Hartmann2021LongHorizonMR}.